\documentclass[letterpaper, 10 pt, conference]{ieeeconf}  % Comment this line out if you need a4paper

\IEEEoverridecommandlockouts                              % This command is only needed if 
\usepackage[utf8]{inputenc} % allow utf-8 input
\usepackage[T1]{fontenc}    % use 8-bit T1 fonts
\usepackage{hyperref}       % hyperlinks
\usepackage{url}            % simple URL typesetting
\usepackage{booktabs}       % professional-quality tables
\usepackage{amsfonts}       % blackboard math symbols
\usepackage{nicefrac}       % compact symbols for 1/2, etc.
\usepackage{microtype}      % microtypography
\usepackage{xcolor}         % colors
\usepackage{graphicx}
\usepackage{multicol}
\usepackage{multirow}
\usepackage{diagbox}

\usepackage{graphics} % for pdf, bitmapped graphics files
\usepackage{amsmath} % assumes amsmath package installed
\usepackage{amssymb}  % assumes amsmath package installed

\usepackage[linesnumbered,ruled,vlined]{algorithm2e}

\title{\LARGE \bf
Polaris: Open-ended Interactive Robotic Manipulation via \\ Syn2Real Visual Grounding and Large Language Models
}

\author{Tianyu Wang$^{1}$, Haitao Lin$^{1}$, Junqiu Yu$^{1}$, Yanwei Fu$^{1 \dagger}$% <-this % stops a space
\thanks{$\dagger$ indicates corresponding author.}
\thanks{$^{1}$Tianyu Wang, Haitao Lin, Junqiu Yu and Yanwei Fu are with Fudan University, China. Corresponding author's email: yanweifu@fudan.edu.cn.}
\thanks{$^{1}$The computations in this research were performed using the CFFF platform of Fudan University.}
}

\begin{document}

\maketitle
\thispagestyle{empty}
\pagestyle{empty}

%%%%%%%%%%%%%%%%%%%%%%%%%%%%%%%%%%%%%%%%%%%%%%%%%%%%%%%%%%%%%%%%%%%%%%%%%%%%%%%%
\begin{abstract}

% It is a longstanding goal for robot to interact with human by understanding the natural language instructions under the open-ended interactive manner.
% robot manipulation using natural language instructions is a long-term goal for human-robot interaction. 
% Although there have been attempts to integrate Large Language Models (LLMs) into robotic systems to capitalize on their powerful task planning capabilities, most have not incorporated visual grounding. 
% This lack of integration means robots are unable to perform feasible manipulations for physical interactions with the environment.
This paper investigates the task of the open-ended interactive robotic manipulation on table-top scenarios. 
While recent Large Language Models (LLMs) enhance robots' comprehension of user instructions, their lack of visual grounding constrains their ability to physically interact with the environment. This is because the robot needs to locate the target object for manipulation within the physical workspace.
To this end, we introduce an interactive robotic manipulation framework called Polaris, which integrates perception and interaction by utilizing GPT-4 alongside grounded vision models.
For precise manipulation, it is essential that such grounded vision models produce detailed object pose for the target object, rather than merely identifying pixels belonging to them in the image. Consequently, we propose a novel Synthetic-to-Real (Syn2Real) pose estimation pipeline. This pipeline utilizes rendered synthetic data for training and is then transferred to real-world manipulation tasks. The real-world performance demonstrates the efficacy of our proposed pipeline and underscores its potential for extension to more general categories. Moreover, real-robot experiments have showcased the impressive performance of our framework in grasping and executing multiple manipulation tasks. This indicates its potential to generalize to scenarios beyond the tabletop. More information and video results are available here: \href{https://star-uu-wang.github.io/Polaris/}{https://star-uu-wang.github.io/Polaris/}.
% Our code and data are available on the project page~\footnote{Project Page.~\url{https://xxx.github.io/Polaris}}. 
% Our code and data are available on the project page~\footnote{Project Page.~\url{https://xxx.github.io/Polaris}}. 

% has demonstrated remarkable results in testing on real-world objects. Moreover, tabletop-level real-robot experiments have showcased the impressive performance of our framework in grasping and executing multiple manipulation tasks. 
% See the project website at: \href{https://star-uu-wang.github.io/WALL-E/}{star-uu-wang.github.io/WALL-E/}
\end{abstract}

%%%%%%%%%%%%%%%%%%%%%%%%%%%%%%%%%%%%%%%%%%%%%%%%%%%%%%%%%%%%%%%%%%%%%%%%%%%%%%%%
\section{Introduction}
% Bridging the interaction between robots and humans to support real-world manipulation tasks is a longstanding goal of robotics research~\cite{billard2019trends}. Natural language instructions are at the core of open-ended human-robot interaction, guiding robot to complete various tasks~\cite{billing2023language, tellex2020robots}. Recently, Large Language Models (LLMs) and Vision Language Models(VLMs) have shown substantial progress~\cite{zhao2023survey}. They possess a wealth of world knowledge and have demonstrated a strong ability to understand human instructions, which have facilitated the emergence of numerous methods for converting language/visual inputs into robotic manipulation actions~\cite{mai2023llm, huang2023voxposer, wu2023tidybot, brohan2023can}. These methods, through their varied attempts across different dimensions of robotics research, prompt us to further contemplate how to fully leverage the perceptual and interactive capabilities of LLMs to support diverse robotic manipulations.

The longstanding goal of robotics research has been to bridge the interaction between robots and humans for real-world grasping~\cite{lin2022know,sun2023language} and manipulation tasks~\cite{billard2019trends,lin2023pourit}. Natural language instructions play a central role in open-ended human-robot interaction, guiding robots to accomplish various tasks~\cite{billing2023language, tellex2020robots,cheang2022learning,sun2023language}. Recently, Large Language Models (LLMs) and Vision Language Models (VLMs) have made significant progress~\cite{zhao2023survey}. They possess extensive world knowledge and have demonstrated strong abilities to understand human instructions, leading to the development of numerous methods for translating language and visual inputs into robotic manipulation actions~\cite{mai2023llm, huang2023voxposer, wu2023tidybot, brohan2023can}. These methods, with their diverse attempts across different dimensions of robotics research, prompt further consideration on how to fully leverage the perceptual and interactive capabilities of LLMs to support various robotic manipulations.

% We investigate the problem of open-ended interactive robotic manipulation (e.g., "Please help me tidy the table"),  within tabletop-level action spaces. Previous work~\cite{huang2023grounded, wake2023chatgpt, mu2024embodiedgpt} has attempted to address such problems by utilizing LLMs as task planners, parsing high-level open-ended instructions into action sequences that they understand. However, for real-world robotic manipulation tasks, existing methods have flaws in visual grounding, and they often do not adequately consider the affordance of objects and the executability of actions. Vision-centric robotic manipulation endows robots with a certain level of environmental perceptual ability, enabling them to plan actions based on perception. However, this comes with the accompanying requirements for high-quality, real-world annotated data.
We explore the issue of open-ended interactive robotic manipulation on tabletop-level scenarios, such as "Please help me tidy the table". Previous studies~\cite{huang2023grounded, wake2023chatgpt, mu2024embodiedgpt} have attempted to tackle such challenges by employing LLMs as task planners, translating high-level instructions into action sequences they comprehend. However, existing methods for real-world robotic manipulation tasks often lack robustness in visual grounding and tend to overlook object affordances and action feasibility. Vision-centric robotic manipulation equips robots with environmental perceptual abilities, enabling action planning based on perception. Nevertheless, this necessitates high-quality, real-world annotated data.

\begin{figure}[t]
    \centering
    \includegraphics[width=0.95\columnwidth]{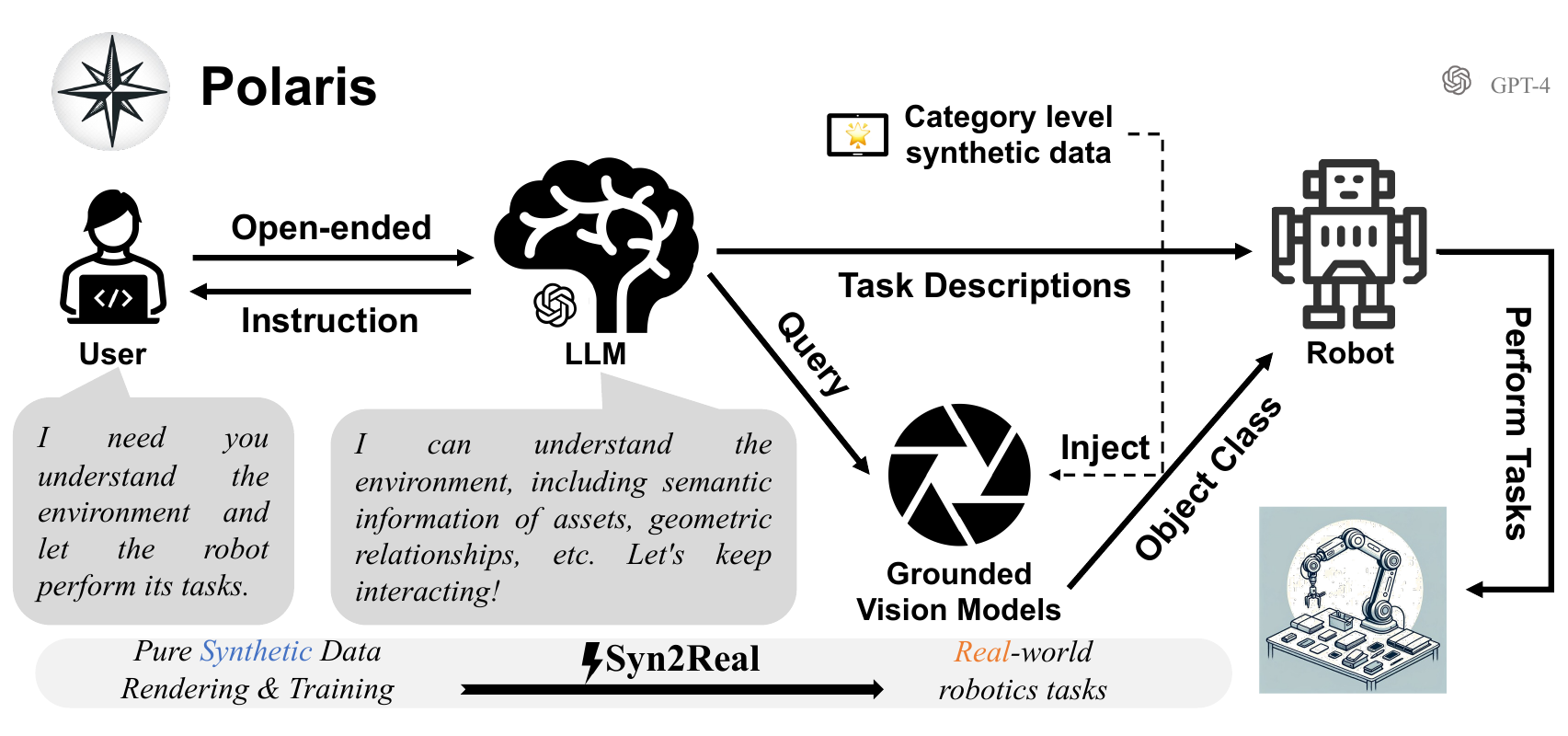}
            \vspace{-0.1in}
    \caption{\textbf{Polaris:} A tabletop-level object robotic manipulation framework centered on syn2real visual grounding driven by open-ended interaction with GPT-4. Users engage in continuous, open-ended interaction with LLM, which maintains an ongoing comprehension of the scenes. 3D synthetic data is integrated into the training of grounded vision modules to facilitate the execution of real-world tabletop-level robotic tasks.\label{fig:teaser}}
        \vspace{-0.25in}
\end{figure}

To tackle the challenge of open-ended interactive robotic manipulation, we use the readily available and powerful Large Language Model (LLM)—GPT-4~\cite{achiam2023gpt} to comprehend and extract the target query from the user's intricate description.
% GPT-4~\cite{achiam2023gpt} has been proven to possess capabilities in image understanding, semantic extraction, and instruction decomposition, making it applicable to a variety of downstream tasks. 
% It also holds considerable potential in robot scene perception and object labelling in text form. 
% Furthermore, we assume that the LLM has understood the user's instructions and provided responses about objects and manipulation methods. 
Once the target query for an object is established, the subsequent step involves the robot locating and grasping the object. Visual grounding enables agents to interpret the visual environment based on these queries, thus aiding in more intricate tasks and interactions~\cite{kim2023gvcci}. Additionally, the 6 Degree-of-Freedom (DoF) object pose estimation serves as a basis for accurate manipulation~\cite{wang2019normalized}. Hence, it is imperative to develop a grounded vision module combined with a pose estimation model to obtain object poses for subsequent motion planning.
% However, a major challenge arises due to the requirement for CAD models during the training phase for both instance-level and category-level pose estimation. This challenge is particularly evident for category-level pose estimation works like NOCS~\cite{wang2019normalized}, which integrate CAD models into real-world scenes to render RGB images and annotate 6D poses. 
A recent method~\cite{wang2019normalized} has extended pose estimation from instance-level to category-level and introduced a category-level dataset with pose annotations. However, this dataset only includes a limited number of categories. To encompass a broader range of categories, we propose an efficient pipeline for generating synthetic data. Leveraging off-the-shelf rendering technologies, we can produce synthetic images of objects with pose annotations.
% we place 3D models into a lightweight rendering engine to 
% generate depth images and generate pose annotations. 
The purely synthetic data generated through rendering is utilized to train the category-level pose estimation model and conduct inference in real-world scenes, representing a novel Synthetic-to-Real (Syn2Real) approach. Ultimately, we seamlessly integrate the vision grounding module with the LLM and the robot planner, establishing an open-ended interactive robot framework.

Our framework, named \textbf{Polaris}, features syn2real visual grounding driven by GPT-4 to enhance tabletop-level interactive robotic manipulation, as depicted in Fig.~\ref{fig:teaser}. Specifically, the framework relies on LLM for scene perception and open-ended human-robot interaction. It trains the pose estimation model within the grounded vision module using purely synthetic data, interprets queries provided by the LLM, and ultimately executes tabletop-level tasks through a 6D pose-based planner, enabling continuous interaction.
% Our experimental results, both qualitative and quantitative, demonstrate the feasibility of this framework for real-world robotic manipulation and its robustness in complex scenarios. Moreover, this approach also has significant potential to generalize to a broader range of scenarios beyond the table.

% Our contributions are summarized as follows:
% \noindent(1) We introduced an automated method for generating depth images and pose annotations when 3D models are available, based on a lightweight rendering engine. Furthermore, we trained MVPoseNet6D using synthetic data and tested the model on real-world images. The results demonstrate that our method achieves syn2real category-level pose estimation and can be effortlessly expanded to a broader range of categories.
% \noindent(2) Based on syn2real visual grounding and GPT-4, we propose a novel framework Polaris, to address the challenge of open-ended interactive robotic manipulation.
% \noindent(3) We demonstrated the capabilities of Polaris through real-robot grasping and manipulation experiments, showing efficient interaction and operational efficiency across a variety of tasks, as well as satisfactory success rates.

Our contributions can be summarized as follows:
(1) We have introduced an automated method for generating depth images and pose annotations when 3D models are available, leveraging a lightweight rendering engine. Additionally, we have trained MVPoseNet6D using synthetic data and evaluated the model on real-world images. The results indicate that our method achieves syn2real category-level pose estimation and can be readily expanded to cover a wider range of categories.
(2) Building upon syn2real visual grounding and GPT-4, we have proposed a novel framework called Polaris to address the challenge of open-ended interactive robotic manipulation.
(3) We demonstrated Polaris's capabilities through real-robot grasping and manipulation experiments, showcasing efficient interaction, operational effectiveness, and satisfactory success rates across various tasks.
% (3) We have demonstrated the capabilities of Polaris through real-robot grasping and manipulation experiments, showcasing efficient interaction and operational effectiveness across various tasks, along with satisfactory success rates.

% contributions

\section{Related Work}
\noindent  \textbf{LLMs for Robotics.}\quad
Embodied intelligence mainly focuses on building systems where agents can purposefully exchange energy and information with the physical environment. It requires a correct understanding of the embodied perception process from a high-dimensional cognitive perspective to a low-dimensional execution perspective~\cite{roy2021machine, romero2023perspective}. Recent work~\cite{mai2023llm} has shown that using LLMs as robotic brain can unify egocentric memory and control by studying downstream tasks of active exploration and embodied question answering. However, such new framework's perception system has flaws in its visual grounding, hindering robot-environment interaction, which will be addressed in this paper. On the other hand, there are zero-shot or few-shot methods~\cite{vemprala2023chatgpt, wake2023chatgpt, brohan2023can, huang2023grounded, wu2023tidybot, mu2024embodiedgpt} that utilize LLMs as task planners, decomposing high-level instructions into executable primitive tasks. These methods assume the ability of robots to execute advanced commands. Unfortunately, they have not yet fully supported the open-ended interaction with robot and not robust enough due to insufficient perception of environment. Instead, our framework  addresses these issues, providing a flexible paradigm that bridges users, LLMs, and robots, offering a new perspective on universal human-robot interaction.

\noindent \textbf{Category-level Object Pose Estimation.}\quad
The 6D object pose estimation is crucial in various applications, such as robotic manipulation and autonomous driving. The objective of category-level object pose estimation is to predict the 6D pose and 3D size of diverse instances belonging to a shared category. The current mainstream methods can be divided into two types: RGB-D based and depth based only. RGB-D based methods ~\cite{wang2019normalized, lin2021dualposenet, wang2021category, lin2022category, chen2021sgpa} often leverages color cues for improved object recognition, which can capture fine-grained texture details, enhancing feature extraction. However, RGB-D based methods often encounter some challenges, such as being sensitive to lighting conditions and color variations. Depth based methods ~\cite{lin2022sar, chen2021fs, di2022gpv, zhang2022rbp, zhang2022ssp} rely solely on depth information, which reduce data complexity and lead to faster processing potentially. 
% In addition, there are some methods that focus on solutions for the specific settings of pose estimation. For example, MobilePose ~\cite{hou2020mobilepose} and CenterPose ~\cite{lin2022single} attempt to reconstruct object pose from a single RGB image. 
% GenPose ~\cite{zhang2023genpose} introduces a novel approach to category-level object pose estimation by leveraging score-based diffusion models with strong generalizability and adaptability. ~\cite{xu2024rnnpose} solves the problem for severe clutter and occlusion in the scenes by recurrent correspondence field estimation. 
The above methods often involve scanning real objects or annotating images of real scenes. Based on the recognition that SAR-Net ~\cite{lin2022sar} is depth based only and the affordance of synthetic data, we opt to further enhance the category-level pose estimation capabilities of SAR-Net and realize real-world application via synthetic-to-real.
Given the framework demands for operational efficiency, our work aims to support a large scale of categories with a minimal number of parameters, thereby establishing a lightweight and easily expandable category-level data rendering and training architecture.

\begin{figure*}[t]
    \centering
    \includegraphics[width=0.9\textwidth]{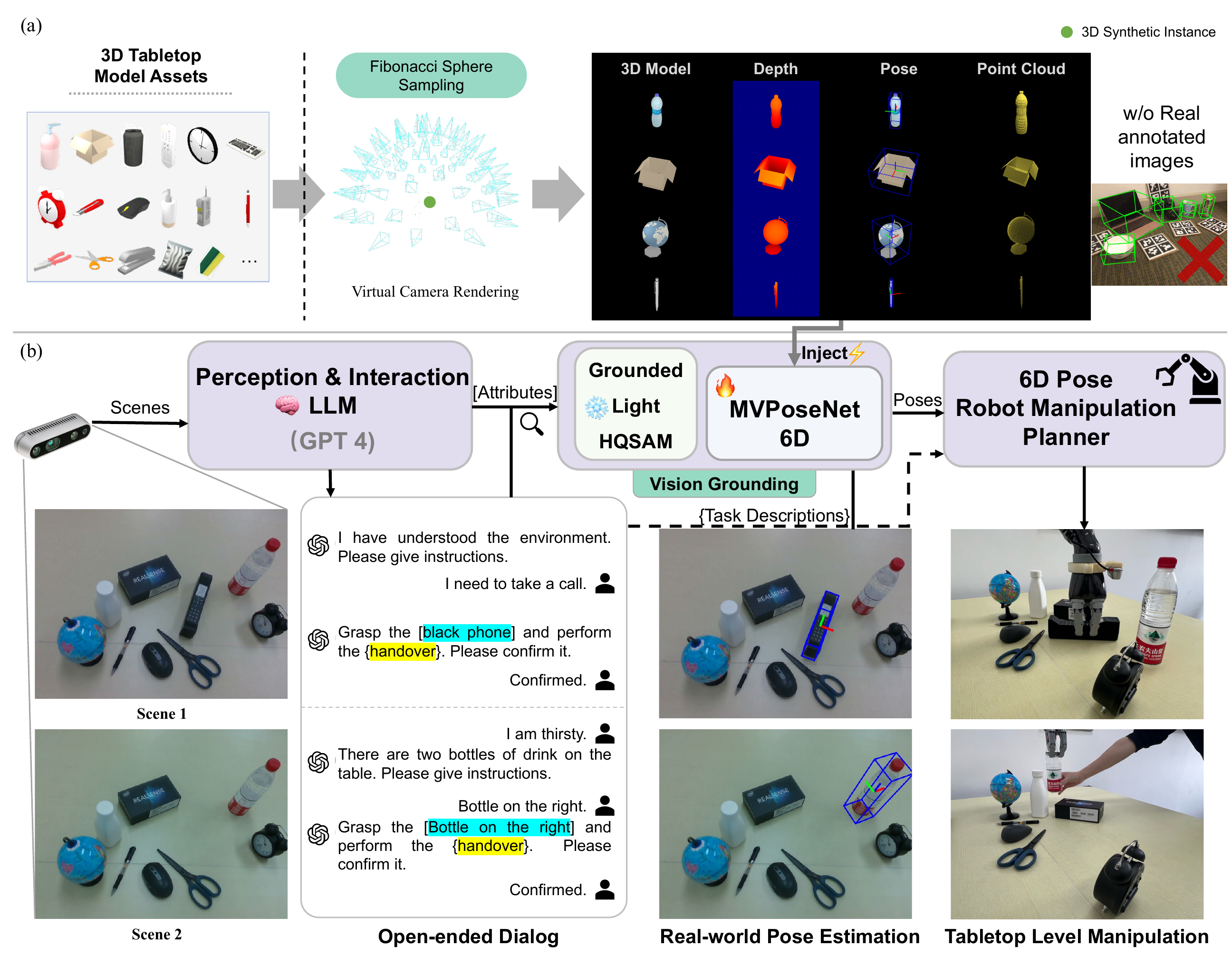}
            \vspace{-0.1in}
    \caption{\textbf{Overview of our framework.} 
    (a) 3D synthetic data rendering. During rendering, we automatically generate various synthetic data by loading 3D model assets into a simulation engine and deploying dynamic virtual camera. We use the Fibonacci Sphere Sampling to select rendering viewpoints, to generate corresponding RGB, depth, pose, and observable point clouds. 
    % (a) 3D synthetic data rendering. During rendering, we automatically generate various attributes of synthetic data by loading 3D model assets into a simulation engine and deploying dynamic virtual camera. We employ the Fibonacci Sphere Sampling to select viewpoints, with rendering corresponding RGB, depth, pose, and observable point clouds. Subsequently, the pure rendered data is integrated into the training of vision module, wherein no real annotated datasets are utilized. 
    (b) The vision-centric robotic task pipeline. 
    Given the image of the scene, which GPT-4, prompted as a scene perception and interaction LLM, interprets to understand instructions and describe objects and tasks. Our parser interprets these descriptions. We freeze the pre-trained detector and segmentation model within the grounded vision models and use a synthetic dataset to train the category-level pose estimation model. After retrieving object attributes, the model predicts poses based on the scene, allowing a 6D pose robot manipulation planner to execute real-world tasks.
    % The scene captured by depth camera is apprehended by GPT-4, which we prompt to serve as a scene perception and interaction LLM. The LLM comprehends open-ended instructions and describes the objects and tasks. A parser we have constructed interprets the attributes and task descriptions. Within the grounded vision models, we freeze the pre-trained detector and segmentation model, and the synthetic dataset rendered in (a) is employed to train the category-level pose estimation model. Once the grounded vision model retrieves object attributes, it predicts the pose based on the current scene. A planner for 6D pose robot manipulation then executes real-world tasks.
    \label{fig:pipeline}}
        \vspace{-0.15in}
\end{figure*}
% This process encompasses the rendering of synthetic datasets, followed by the training of vision model exclusively with pure synthetic data.

\noindent \textbf{Vision-centric Interactive Robot Manipulation.}\quad
In the realm of vision-centric interactive robot manipulation, recent advancements focus on enhancing robots' ability to perform tasks by learning from human demonstrations and integrating LLMs or Large Multimodal Models (LMMs) for better understanding and execution of vision-centric tasks. Interactive robot manipulation learning from human demonstrations frequently demands high-quality human videos or teleoperation data~\cite{xiong2021learning, shao2021concept2robot, chen2021learning, sontakke2024roboclip, cui2022play, wang2023mimicplay, spisak2024robotic}. Simultaneously, it requires dependable reinforcement learning or imitation learning algorithms for the training of robot policies~\cite{englert2018learning, zhu2023viola, florence2022implicit, zhang2018deep, kumar2023graph, xu2023xskill}. While these methods offer considerable flexibility, they all necessitate the collection of human demonstrations through various means to learn different tasks, often requiring real-world physical annotations~\cite{brohan2023rt, padalkar2023open}. Interactive robotic manipulation frequently requires affordance learning for objects based on visual inputs, where zero-shot~\cite{huang2023voxposer, zhang2023large}, few-shot~\cite{wang2021demograsp, chao2023fewsol}, and open-ended learning~\cite{kasaei2019interactive, yang2022interactive, ayoobi2023explain} are of significant interest. Open-ended learning methods facilitate the update and expansion of category sets and also provide a broader interaction space for human-in-loop tasks. Our proposed model employs a convenient and efficient method to render synthetic data for training the pose estimation models within the grounded vision module. By integrating with GPT-4, it addresses object affordance and supports open-ended human-robot interaction.
% The entire pipeline exhibits commendable continuous learning capabilities and scalability.
% \subsection{Selecting a Template (Heading 2)}

% First, confirm that you have the correct template for your paper size. This template has been tailored for output on the US-letter paper size. 
% It may be used for A4 paper size if the paper size setting is suitably modified.

% \subsection{Maintaining the Integrity of the Specifications}

% The template is used to format your paper and style the text. All margins, column widths, line spaces, and text fonts are prescribed; please do not alter them. You may note peculiarities. For example, the head margin in this template measures proportionately more than is customary. This measurement and others are deliberate, using specifications that anticipate your paper as one part of the entire proceedings, and not as an independent document. Please do not revise any of the current designations

\section{Problem Formulation}
We present a novel open-ended interactive robotic manipulation problem via syn2real visual grounding and LLMs here, which shall have the following desired properties.

\noindent  \textbf{Property~1.} 3D synthetic data rendering: For arbitrary 3D model assets, categorized by object class, synthetic instance data is needed to be collected from various viewpoints through a virtual engine and then injected into the training of subsequent category-level pose estimation model.

\noindent  \textbf{Property~2.} Open-ended interactive robotic manipulation: Based on scene inputs acquired from RGB-D camera, the prompted LLM must comprehend the scene and understand the user's natural language instructions, labelling the target objects for interaction and specific task descriptions. The robot must grasp and execute tasks according to the 6D pose of the target objects.

In terms of above descriptions, we give the typical setups to validate our framework:
\textbf{(i)} We are particularly interested in  table-level robotic tasks, where the variety of reachable objects is confined within an almost known domain. Hence, we opt to render common table objects category data for training the pose estimation model. However, based on our proposed rendering method, it is feasible to collect additional synthetic pose data from existing datasets (~\cite{chang2015shapenet, Mo_2019_CVPR}, etc) or custom 3D modeling data, which can be generalized to the training of pose estimation models with greater capacity.
\textbf{(ii)} Furthermore, we prefer the manner of interactive instructions. Specifically, user natural language instructions may not directly specify the objects for interaction (some queries may even be ambiguous), necessitating understanding and labelling by the LLMs. Additionally, continuous interaction is required, with the LLMs needing to keep up with scene changes and next user instructions.
%
% In the definition of the aforementioned problem, the following assumptions will also be adhered to:
% *
%
%\noindent  \textbf{Assumption 1.} 

% Note that, we advocate that our Polaris shall be typically utilized to address the scenarios above. Essentially,  our system has the potential of generalizing the scenarios beyond above setups, which however, is beyond the scope of this paper.

% To resolve open-ended interactive robotic manipulation problem, we equip the robotic grasping system with a framework called Polaris. 

% tabletop-level
\section{Method}
Polaris is a sophisticated interactive robotic manipulation framework integrating perception and interaction, employing a LLM, specifically GPT-4, with grounded vision models. An overview of the proposed Polaris framework is presented in Fig.~\ref{fig:pipeline}. In the ensuing subsections, we will detail the synthetic data rendering (Sec.~\ref{sub:synthetic data rendering}), synthetic-to-real category-level pose estimation (Sec.~\ref{sub:syn2real category-level pose estimation}, and open-ended interactive robotic manipulation design (Sec.~\ref{sub:open-ended interactive robotic manipulation design}).

\subsection{3D Synthetic Data Rendering} \label{sub:synthetic data rendering}
Given a 3D model $M$ from the model collections $\mathcal{M}, M \in \mathcal{M}$, we aim to render the RGB images $I$, depth image $D$, partial point cloud $P$ and calculate the 6-DoF pose transformation $T=(R,t)$ and 3D size $s$ of the model from current camera viewpoints.

Leveraging the SAPIEN~\cite{Xiang_2020_SAPIEN} simulation environment, we utilize a subset of 3D models from the PartNet-Mobility dataset~\cite{Mo_2019_CVPR}, supplemented with custom CAD modeling data. To acquire the rendered images for each 3D model, we position the model at the origin of the world frame, variously adjusting the camera viewpoint to capture and render corresponding depth images. The PartNet-Mobility dataset comprises 2,000 articulated objects with motion annotations and rendering materials. This dataset serves as a valuable resource for advancing research in generalizable computer vision and manipulation, representing a continuation of the pioneering work in ShapeNet and PartNet.

In particular, to capture a broader range of camera viewpoints, we employ Fibonacci sphere sampling method to evenly distribute the camera positions across a sphere, as shown in Fig.~\ref{fig:pipeline} (a). Additionally, we introduce random in-plane rotations to each camera's orientation, expanding the coverage to encompass a more diverse set of camera angles.

Ultimately, we rendered a total of 24 tabletop-level object classes, including \{"Bottle", "Box", "Dispenser", "Remote", "Camera", "Clock", "Eyeglasses", "Fan", "Faucet", "Globe", "Kettle", "Keyboard", "Knife", "Lamp", "Laptop", "Mouse", "Pen", "Phone", "Pliers", "Scissors", "Stapler", "USB", "Packaging", "Sponge"\}, with 1K instances, resulting in 300K depth images along with poses, as illustrated in Fig.~\ref{fig:pipeline} (a). Additionally, the corresponding RGB and point cloud were generated simultaneously. This stage of the process solely relied on the CPU, making it very efficient. The pseudocode for the rendering process is provided by Algorithm~\ref{alg1}.

\vspace{-0.1in}
\begin{algorithm}
\footnotesize
\SetKwInput{KwData}{Input}
\SetKwInput{KwResult}{Output}
\caption{Synthetic Data Rendering}
\label{alg1}
\KwData{3D models $I$ containing $N$ categories}
\KwResult{Fibonacci sphere rendering data for instances}
Initialization: Set the rendering engine and parameters

\For{$i\leftarrow 1$ \KwTo $N$}{
    % w instances for the class
    \emph{The number of instances $W$ contained in class $i$}\;
    \For{$j\leftarrow 1$ \KwTo $W$}{
    Load the URDF model $U_j$ of instance $j$\;
    $(X_{min}, Y_{min}, Z_{min})\leftarrow \infty$, $(X_{max}, Y_{max}, Z_{max})\leftarrow -\infty$

    \emph{The number of parts $S$ contained in model $U_j$}\;
        \For{$k\leftarrow 1$ \KwTo $S$}{
        Load the points $P_{j}^k$ of the part $k$\;
        $(x_{min}, y_{min}, z_{min})\leftarrow {P_{j}^k}_{min}$, $(x_{max}, y_{max}, z_{max})\leftarrow {P_{j}^k}_{max}$
        
        Update global extreme point $(X_{min}, Y_{min}, Z_{min})$ and $(X_{max}, Y_{max}, Z_{max})$\;
        }
    Compute scale $S_j$ by $(S_{j}^X, S_{j}^Y, S_{j}^Z)\leftarrow (X_{max} - X_{min}, Y_{max} - Y_{min}, Z_{max} - Z_{min})$\;
    Generate camera poses $\tau$ by \emph{Sphere Sampling}\;
    \For{$n, {\tau}_n\leftarrow enumerate (\tau)$}{
    Mount dynamic virtual camera ${\tau}_n$\;
    Get instance pose ${\lambda}_{j}^n\leftarrow {{\tau}_n}^{-1}$\;
    Update render to get RGB, PointCloud and Depth under ${\tau}_n$\;
    }
    }
}
% \While{condition}{
%     Do something\;
%     \If{condition}{
%         Do something else\;
%     }
% }
\end{algorithm}
\vspace{-0.15in}

\subsection{Syn2Real Category-level Pose Estimation} \label{sub:syn2real category-level pose estimation}
Considering the efficient runtime requirements of the robotic manipulation, we aim to support a greater number of object categories with a smaller number of parameters and to robustly facilitate pose estimation in tabletop scenarios under various lighting conditions. We extend the output dimension of the original decoder in the depth-only SAR-Net~\cite{lin2022sar} from 6 to 24 to accommodate the 24 new categories. By utilizing synthetically rendered multi-view data, the training process remains consistent with the original SAR-Net. This enables the model to learn shared geometric features among intra-category instances from different views of observed shapes.
% We further extend the encoder dimension of the depth-only SAR-Net~\cite{lin2022sar}, injecting multi-view synthetic data into the training module, resulting in the Multi-View (MVPoseNet6D). 
For the processing of category-level templates, we randomly select a general instance within the class, perform Object Canonicalization to align the coordinate system, and then execute Poisson Sampling and Farthest Point Sampling (FPS) to extract the category-level template point cloud. We transfer the model trained on synthetic data to the inference module, to support real-world category-level object pose estimation.

\subsection{Open-ended Interactive Robotic Manipulation} \label{sub:open-ended interactive robotic manipulation design}
As shown in Fig.~\ref{fig:pipeline} (b), our open-ended robot interaction framework primarily comprises three modules: a LLM that supports scene perception and human-robot interaction, a vision grounding module, and a robotic manipulation planner based on the 6D pose of objects.

\noindent\textbf{Perception and Interaction LLM.}\quad
Based on scene inputs from a depth camera, the framework we aim to construct should be capable of perceiving the scene, identifying object assets on the tabletop, and engaging in continuous interaction with the user based on the affordance of these assets and user requirements. By leveraging GPT-4's capabilities in image understanding, semantic extraction, and its powerful ability to comprehend user instructions, we call the GPT-4 API and prompt it to serve as the high-level perception and interaction brain for the robot. Firstly, we provide GPT-4 with a system-level explanation. This explanation is designed to guide the LLM to affirm its role and confine it within a specific domain, ensuring robustness and professionalism in task analysis during robotic manipulation. The primary task of the LLM before interacting with users is to understand and learn about the spatial specifics of tasks, rather than initiating work directly. Simultaneously, we expect the responses from the LLM to be task-oriented, necessitating specific object queries and task descriptions. We have constructed a parser that, upon the LLM's comprehension of the user's intent during the $i$ round of interaction and subsequent user confirmation of the robotic instructions, extracts the object attributes $A_i$ and task descriptions $T_i$ from the instructions. These are then passed on to the vision grounding module and the robot planner.

\noindent\textbf{Visual Grounding.}\quad
The vision grounding module, serving as the core of vision-centric robotic manipulation, receives raw RGB images of the scene captured by depth cameras, depth information, and attributes $A_i$ derived from the parsing of instruction provided by the LLM. We treat the attributes $A_i$ as a query, which is then sent to the frozen Grounded-Light-HQSAM. Grounded-Light-HQSAM is an integrated model that incorporates Grounding DINO~\cite{liu2023grounding} and HQ-SAM~\cite{sam_hq}. Grounding DINO functions as an open-set object detector, utilizing visual-language modality fusion to generate bounding boxes and labels with free-form referring expressions. This process involves multiple phases, including a feature enhancer, a language-guided query selection module, and a cross-modality decoder. Once the grounding box of the target object is obtained, it is used as a prompt for the segmentation model. Grounded-Light-HQSAM is capable of generating refined object masks in a lightweight and relatively fast manner. These masks are then used to crop the depth pixels belonging to the object, facilitating follow up pose and size estimation. After obtaining the depth, grounding mask and category label of the target object, we use the trained MVPoseNet6D to recover the real-world object 6-DoF pose and 3D size in the camera coordinate system, which provides information about the object's state in the current scene. 
Then, the estimated pose is transformed into the robot's base coordinate system for further motion planning.

\noindent\textbf{6D Pose Robot Manipulation Planner.}\quad
First of all, we need to answer a question: \textit{Why is it necessary to consider the object's pose instead of using a straightforward grasp pose estimation method?} - While methods for direct 6-DoF grasp pose estimation~\cite{mousavian20196, fang2020graspnet, fang2022anygrasp} have made significant progress, their scope remains limited and is primarily applicable to pick-and-place operations. In our framework, we integrate pose estimation methods because the state of objects in 3D space is crucial for calculating meaningful manipulation points in various operational tasks, such as pouring water, handover, and some compositional tasks. We believe that object pose provides robots with rich contextual knowledge before proceeding with the motion planning. To ensure the extensibility of Polaris, we follow the principle of first inferring the object's pose and then calculating useful target gripper poses for precise manipulations. Particularly, as the pose of intra-category instances are pre-canonicalized~\cite{wang2019normalized}, it becomes advantageous to define category-level grasp poses relevant to each task. These defined grasp poses are then transformed from object coordinates to camera coordinates using the estimated object 6D pose. The robot executes motion planning to move its gripper to the target grasp pose to complete each task. Thus, we have constructed a task-oriented 6D Pose Robot Manipulation Planner.
% By understanding common tabletop-level object classes and manipulation tasks, and utilizing the 6D pose of objects along with the underlying ROS control for robots, we have constructed a task-oriented 6D Pose Robot Manipulation Planner.

\section{Experiment}
We conducted a series of experiments, which included evaluating the synthetic-to-real pose estimation in both single-object and multi-object real-world scenarios, as well as testing the proposed Polaris framework against several baseline methods. The goals of the experiments are 1) to investigate the feasibility of applying MVPoseNet6D, trained purely on synthetic data, in real-world applications; 2) to demonstrate that our Polaris can efficiently achieve elaborate human-robot interaction in various scenarios; 3) to show the accuracy of our tabletop-level robotic manipulation system.

\begin{figure}[h]
    \centering
    \includegraphics[width=0.9\columnwidth]{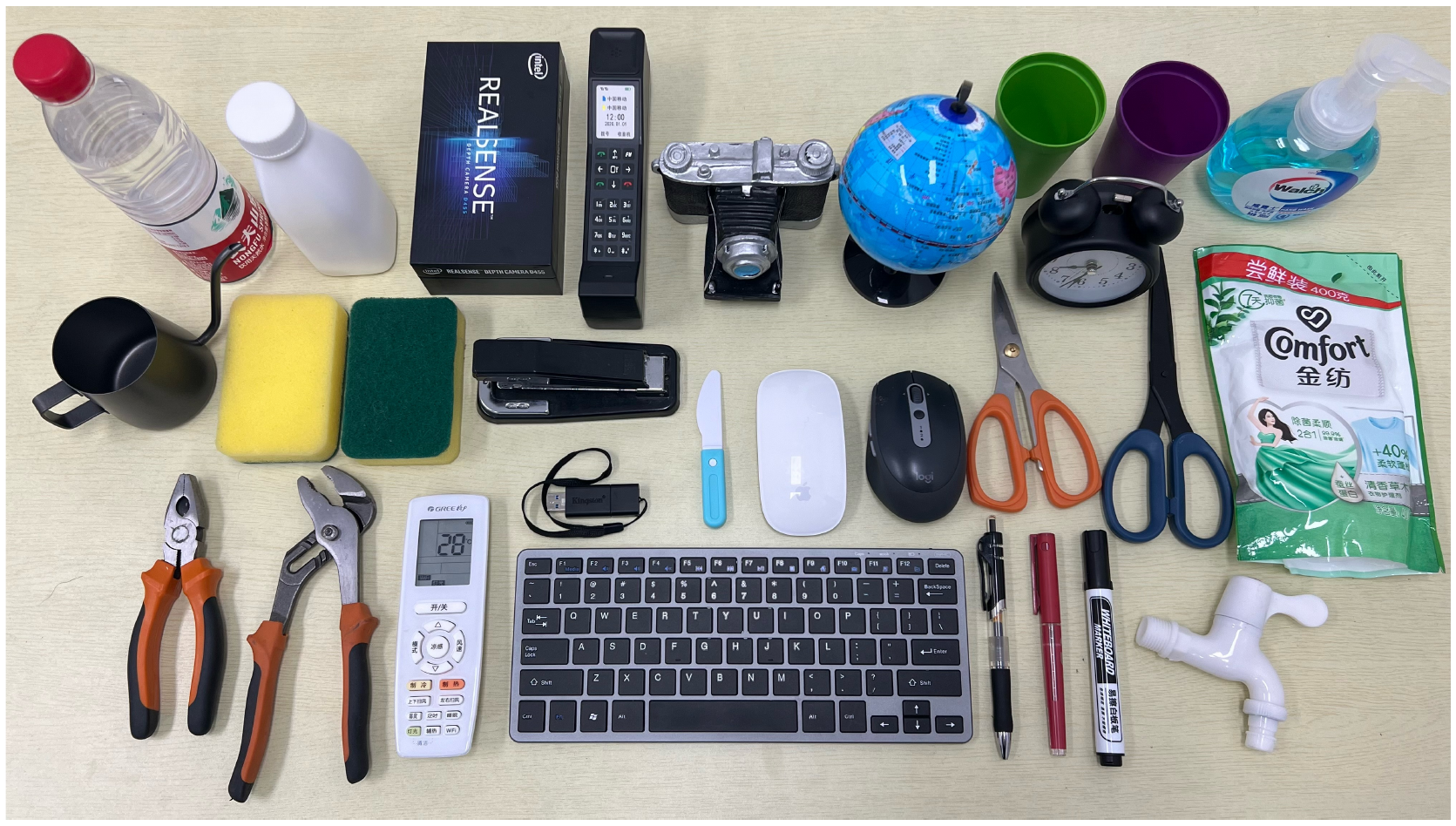}
            \vspace{-0.1in}
    \caption{\textbf{Real-world experimental objects.} We test our method using different instances from multiple tabletop-level objects, some of which are confusing in terms of color, shape, rigidity, deformability, and functionality.\label{fig:test_objects}}
        \vspace{-0.15in}
\end{figure}

Polaris was deployed on a PC workstation with an Intel i9-13900K CPU and an NVIDIA RTX 6000 Ada Generation GPU. We used a KINOVA GEN2 robot with a Realsense D435 depth camera mounted in an eye-in-hand configuration. The tabletop-level testing objects are shown in Fig.~\ref{fig:test_objects}.

\begin{figure*}[htbp]
    \centering
    \includegraphics[width=1.0\linewidth]{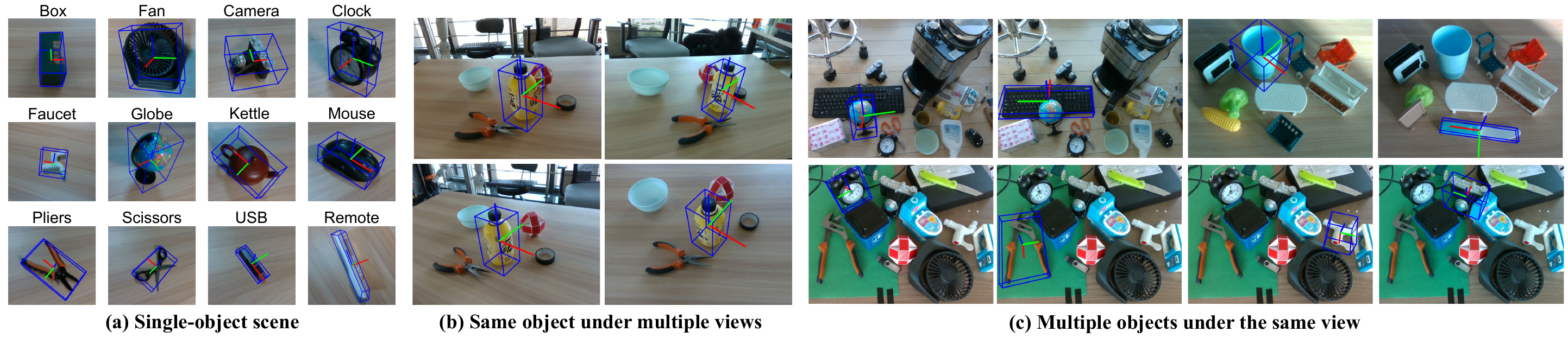}
            \vspace{-0.25in}
    \caption{\textbf{Results of real-world object pose estimation.} (a) Test results of single-object scene. We present a subset of the visualization results of the pose and size estimation using the trained MVPoseNet6D model. The outcomes are represented with a tightly oriented 3D bounding box and colored XYZ-axis. (b) The scene with same object under multiple views. We show the pose of a bottle under different views. (c) The scene with multiple objects under the same view. We show the pose estimation of different objects in several cluttered scenes.\label{fig:pose_estimation}}
        \vspace{-0.1in}
\end{figure*}

\subsection{Real-world Object Pose Estimation Evaluation}
Synthetic-to-real genaralizability is crucial for models trained solely on synthetic data, and the accuracy of pose estimation is foundational for vision-centric interactive robotic manipulation.. Therefore, we visually represent our predictions by displaying the predicted 6D pose and 3D size in the form of a tight-oriented bounding box, as in Fig.~\ref{fig:pose_estimation}.

\begin{table*}[htb]
\centering
\caption{Results of Instance-oriented Grasping}
\label{tab:grasping}
\vspace{-0.1in}
\begin{tabular}{cccccc}
\toprule[1pt]
Method &Accuracy(\%) &\# Questions&Time(ms)&Success / Trials & Success Rate(\%) \\ \midrule[0.5pt]
RandomGrasp & * & * & 43.7 & 22 / 60 & 36.67 \\ \midrule[0.5pt]
\textbf{Ours (full model)} & \textbf{93.52} & \textbf{1.36} & 509.6 & 55 / 60 & \textbf{91.67} \\ 
Ours (w/o 3DGCN) & 90.07 & 1.41 & 442.7 & 53 / 60 & 88.33 \\
Ours (w/o Light-HQSAM | w/ SAM) & 87.92 & 1.97 & 849.6 & 49 / 60 & 81.67 \\
Ours (w/o GPT-4 | w/ GPT-3.5-turbo) & 93.37 & 3.82 & 552.3 & 55 / 60 & 91.67 \\ \midrule[0.5pt]
Ours (w/o FSP | w/ FHemiSP) & 82.69 & 1.39 & 497.8 & 42 / 60 & 70.00 \\
\bottomrule[1pt]
\end{tabular}
\end{table*}

\begin{table*}[htb]
\centering
\caption{Results of Task-oriented Manipulation\label{tab:manipulation}}
\vspace{-0.1in}
\begin{tabular}{ccccccccc}
\toprule[1pt]
\multirow{2}{*}{Method} & \multicolumn{2}{c}{Single-object Scene} & \multirow{2}{*}{Success(\%)} & \multicolumn{2}{c}{Multi-object Scene} & \multirow{2}{*}{Success(\%)} & \multirow{2}{*}{Compositional Tasks} & \multirow{2}{*}{Total Success(\%)} \\ \cmidrule(r){2-3} \cmidrule(r){5-6} 
 & Pick-and-Place & Handover &  & Stack & Tidy &  &  & \\ \midrule[0.5pt]
Ours & 18 / 20 & 17 / 20 & 87.50 & 10 / 15 & 12 / 15 & 73.33 & 6 / 10 & \textbf{78.75} \\ 
\bottomrule[1pt]
\end{tabular}
\vspace{-0.1in}
\end{table*}

To confirm the effectiveness of our model in real-world scenarios, we evaluated instances of 24 predefined tabletop-level categories within a single scene. For each instance, we captured real-world images from various viewpoints. As illustrated in Fig.~\ref{fig:pose_estimation} (a), we visually present a subset of the pose estimation results. The results are depicted using tightly oriented bounding boxes. The object is visibly positioned within and aligned with the box, showing the model's accurate estimation performance. Since the real-world test instances are not used to train the model, these results demonstrate the synthetic-to-real capability of our data rendering method and the model. Furthermore, the examples given in Fig.~\ref{fig:pose_estimation} (b) indicate that the model is capable of consistently generating accurate pose and size results, even with significant changes in viewpoint. To thoroughly evaluate the model's performance in cluttered environments with diverse backgrounds and objects, we place the target instance in a scene with numerous objects. The multi-object scene poses more challenges, as the objects are randomly placed, and some objects occlude each other. This setting allows us to assess the robustness of our method in real-world environments. As depicted in Fig.~\ref{fig:pose_estimation} (c), some objects are partially occluded, but the predicted pose and size remain accurate. These results demonstrate the effectiveness of our method and prove that using our approach allows for a fast and scalable extension of the pose estimator to multiple categories, enabling adaptability to a wider range of objects.

\begin{figure*}[htb]
    \centering
    \includegraphics[width=0.9\linewidth]{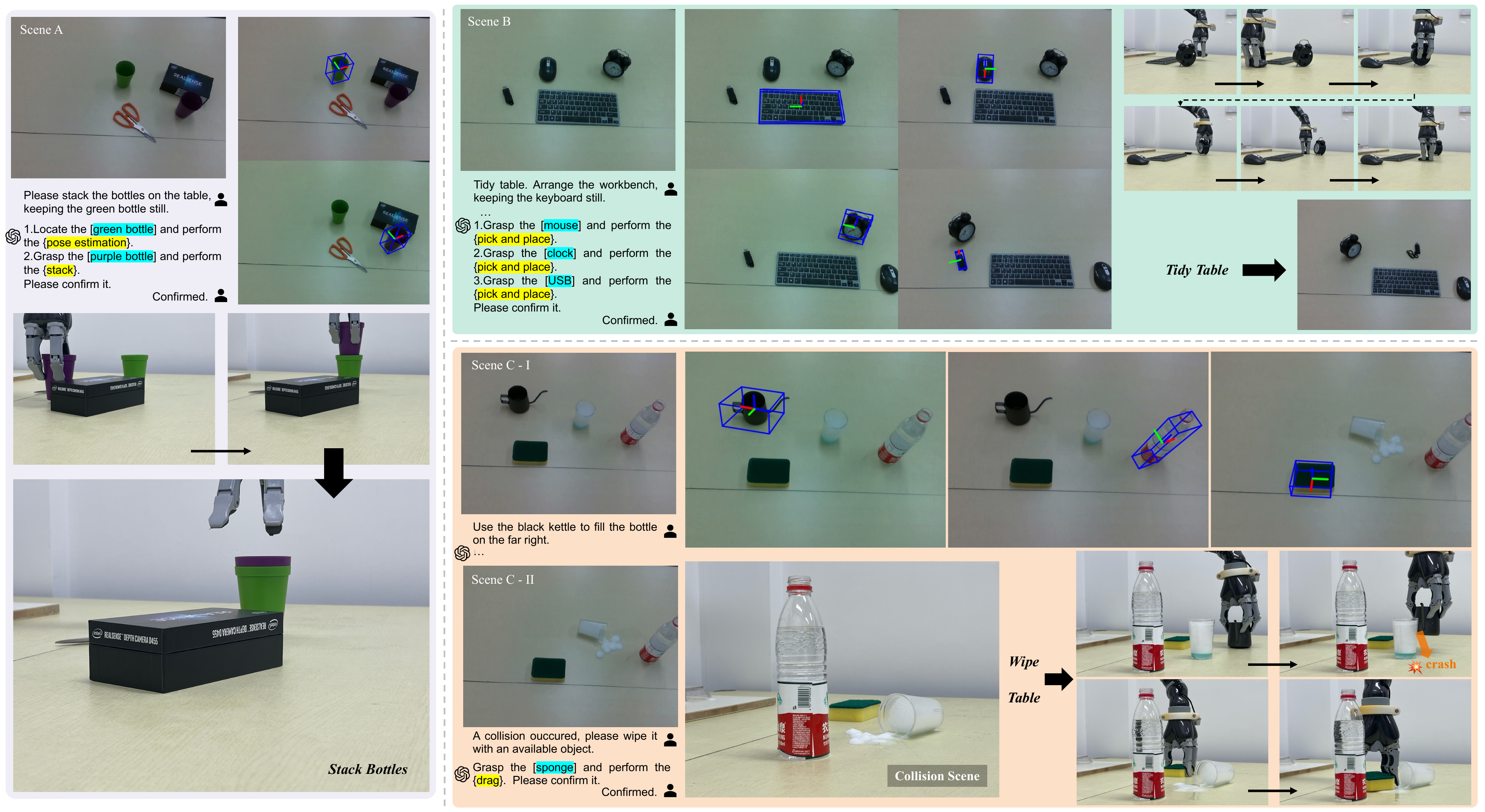}
            \vspace{-0.1in}
    \caption{\textbf{Examples of open-ended interactive real-robot experiments.} Manipulation tasks for three different base scenes are presented, including excerpts from the interaction process between the user and the LLM, the pose estimation results of the manipulated objects in different scenes, and  the keyframes of the robot manipulation. \textbf{Scene A:} Stack bottles on the table. \textbf{Scene B:} Tidy the items of workbench. \textbf{Scene C:} A compositional task considering the affordance of objects after a sudden collision.\label{fig:examples}}
        \vspace{-0.15in}
\end{figure*}

\subsection{Open-ended Interactive Real-Robot Experiments}
The open-ended interactive robot experiments mainly consist of two parts: instance-oriented grasping and task-oriented manipulation. Different methods and tasks share the same experimental environment and hardware.

\noindent\textbf{Instance-oriented Grasping.}\quad
We conducted instance-oriented grasping experiments using the Polaris framework and the following constructed baselines: 1) \textbf{RandomGrasp} randomly selects a grasping target from the instance space generated by the Polaris vision grounding module until it grasps the object requested by the user. 2) \textbf{Polaris (w/o 3DGCN)} omits the trained 3DGCN~\cite{lin2020convolution} used in MVPoseNet6D, where the primary function of 3DGCN is to filter out speckle and background noise of the different category-level objects point cloud captured by the depth camera. 3) \textbf{Polaris (w/o Light-HQSAM | w/ SAM)} replaces Light-HQSAM with the original pre-trained SAM~\cite{kirillov2023segany}. 4) \textbf{Polaris (w/o GPT-4 | w/ GPT-3.5-turbo)} replaces GPT-4 with GPT-3.5-turbo~\cite{wang2023wall}, and the environment's assets are provided by both the labelling model and manually by the user. The task-level prompts for the LLM(GPT-3.5-turbo) are more verbose and complex. 5) \textbf{Polaris (w/o FSP | w/ FSemiSP)} replaces Fibonacci sphere sampling with Fibonacci hemisphere sampling during synthetic data rendering. In Table~\ref{tab:grasping}, we report the runtime, the number of questions asked, runtime, the visual accuracy (calculate the deviation between the estimated 6D pose of real-world objects and manually annotated poses), and the number of successful attempts for each method. First of all, the full model of Polaris achieves a visual accuracy of 93.52\%, which demonstrates the feasibility and effectiveness of our proposed syn2real pose estimation method. The substantial increase in grasping success rate compared to RandomGrasp (from 36.67\% to 91.67\%) validates that the integration of our prompted LLM and vision grounding can help understand user intentions with efficient interaction states (averaging only 1.36 questions per successful grasp). The 3DGCN is trained for the classes used in our experiments to further optimize point clouds, as seen by the improvement in visual accuracy in our table scenes with plain backgrounds. The replacement of the segmentation model demonstrates that our integration of Light-HQSAM significantly reduces runtime (from 849.6ms to 509.6ms), enabling higher grounded vision accuracy at a faster operational efficiency. Furthermore, we conducted an ablation analysis on the synthetic data rendering method, where the visual accuracy obtained from hemisphere sampling decreased by about 10 percentage points. Fibonacci Sphere sampling supported a wider coverage of viewpoints, enabling the handling of more generalized scenes in real-world testing.

\noindent\textbf{Task-oriented Manipulation.}\quad
The high visual accuracy and grasping success rates provide a strong guarantee for task-oriented manipulation. To assess Polaris's capability in robotic manipulation, we evaluated its performance on single-object, multi-object, and compositional tasks, as shown in Table~\ref{tab:manipulation}. Quantitative results indicate that our method performs well in task-oriented manipulation, achieving success rates of 87.5\% for single-object performance and 73.33\% for multi-object performance, with an overall success rate of 78.75\% across all tasks. Simultaneously, a qualitative analysis of the three specific examples of open-ended interactive robot manipulation in Fig.~\ref{fig:examples} reveals that our method successfully accomplished real-world tasks and effectively supported continuous interaction.

\section{Conclusion}
% In this work, we introduce Polaris, a novel framework for open-ended interactive robotic manipulation based on LLMs and syn2real visual grounding. The syn2real pose estimation method, trained with synthetic data, has shown commendable performance on real-world testing. Additionally, both qualitative and quantitative results from tabletop-level real-robot experiments validate the effectiveness of Polaris. We anticipate that Polaris will unleash great potential for generalization across a broader range of complex scenarios in robot manipulation.

This paper introduces Polaris, a novel framework for open-ended interactive robotic manipulation based on LLMs and syn2real visual grounding. Our syn2real pose estimation method, trained with synthetic data, performs well in real-world tests. Tabletop-level real-robot experiments provide validation of Polaris's effectiveness. We anticipate that Polaris will significantly enhance generalization across diverse, complex robotic manipulation scenarios.
% \section{Acknowledgements}
% The computations in this research were performed using the CFFF platform of Fudan University.

% \addtolength{\textheight}{-12cm}   % This command serves to balance the column lengths
                                  % on the last page of the document manually. It shortens
                                  % the textheight of the last page by a suitable amount.
                                  % This command does not take effect until the next page
                                  % so it should come on the page before the last. Make
                                  % sure that you do not shorten the textheight too much.

%%%%%%%%%%%%%%%%%%%%%%%%%%%%%%%%%%%%%%%%%%%%%%%%%%%%%%%%%%%%%%%%%%%%%%%%%%%%%%%%

%%%%%%%%%%%%%%%%%%%%%%%%%%%%%%%%%%%%%%%%%%%%%%%%%%%%%%%%%%%%%%%%%%%%%%%%%%%%%%%%

%%%%%%%%%%%%%%%%%%%%%%%%%%%%%%%%%%%%%%%%%%%%%%%%%%%%%%%%%%%%%%%%%%%%%%%%%%%%%%%%
% \section*{APPENDIX}

% Appendixes should appear before the acknowledgment.

% \section*{ACKNOWLEDGMENT}

% The preferred spelling of the word ÒacknowledgmentÓ in America is without an ÒeÓ after the ÒgÓ. Avoid the stilted expression, ÒOne of us (R. B. G.) thanks . . .Ó  Instead, try ÒR. B. G. thanksÓ. Put sponsor acknowledgments in the unnumbered footnote on the first page.

%%%%%%%%%%%%%%%%%%%%%%%%%%%%%%%%%%%%%%%%%%%%%%%%%%%%%%%%%%%%%%%%%%%%%%%%%%%%%%%%

% References are important to the reader; therefore, each citation must be complete and correct. If at all possible, references should be commonly available publications.

% \clearpage
% \bibliographystyle{IEEEtran.bst}
% \bibliography{egbib}

\end{document}